\documentclass[letterpaper, 10 pt, conference]{ieeeconf}  

\IEEEoverridecommandlockouts                              

\usepackage{graphicx}
\usepackage{booktabs} 
\usepackage{multirow} 
\usepackage{bbm} 
\usepackage{amssymb}
\usepackage{wrapfig} 
\usepackage{array}
\usepackage{graphicx}
\usepackage{subcaption}
\usepackage{marvosym}
\usepackage[marginal]{footmisc}

\usepackage{bm}
\usepackage{amsmath}
\usepackage{hyperref}
\usepackage{grffile}
\usepackage{pifont} 
\usepackage{mathtools} 
\usepackage{tabularx}
\usepackage{float}
\usepackage{booktabs} 
\usepackage{pifont} 
\usepackage{color}

\usepackage[table]{xcolor}

\title{\LARGE \bf
Polyp-Gen: Realistic and Diverse Polyp Image Generation for Endoscopic Dataset Expansion
}

\author{Shengyuan Liu$^{1,\dag}$, Zhen Chen$^{2,\dag}$, Qiushi Yang$^{3}$, Weihao Yu$^{1}$, Di Dong$^{4}$, Jiancong Hu$^{5}$, and Yixuan Yuan$^{1,*}$
\thanks{$^{1}$ S. Liu, W. Yu, and Y. Yuan are with the Department of Electronic Engineering, Chinese University of Hong Kong, Hong Kong SAR, China.}%
\thanks{$^{2}$ Z. Chen is with the Centre for Artificial Intelligence and Robotics (CAIR),
Hong Kong Institute of Science \& Innovation, Chinese Academy of Sciences, Hong Kong SAR, China.}%
\thanks{$^{3}$ Q. Yang is with the Department of Electrical Engineering, The City University of Hong Kong, Hong Kong SAR, China.}%
\thanks{$^{4}$ D. Dong is with the Institute of Automation, Chinese Academy of Sciences, Beijing, China.}
\thanks{$^{5}$ J. Hu is with The Sixth Affiliated Hospital, Sun Yat-sen University, Guangzhou, China.}
\thanks{$\dag$\,Equal contribution, $*$\,Corresponding author.}
}
\begin{document}

\maketitle
\thispagestyle{empty}
\pagestyle{empty}


\begin{abstract}
Automated diagnostic systems (ADS) have shown significant potential in the early detection of polyps during endoscopic examinations, thereby reducing the incidence of colorectal cancer. However, due to high annotation costs and strict privacy concerns, acquiring high-quality endoscopic images poses a considerable challenge in the development of ADS. Despite recent advancements in generating synthetic images for dataset expansion, existing endoscopic image generation algorithms failed to accurately generate the details of polyp boundary regions and typically required medical priors to specify plausible locations and shapes of polyps, which limited the realism and diversity of the generated images. To address these limitations, we present Polyp-Gen, the first full-automatic diffusion-based endoscopic image generation framework. Specifically, we devise a spatial-aware diffusion training scheme with a lesion-guided loss to enhance the structural context of polyp boundary regions. Moreover, to capture medical priors for the localization of potential polyp areas, we introduce a hierarchical retrieval-based sampling strategy to match similar fine-grained spatial features. In this way, our Polyp-Gen can generate realistic and diverse endoscopic images for building reliable ADS. Extensive experiments demonstrate the state-of-the-art generation quality, and the synthetic images can improve the downstream polyp detection task. Additionally, our Polyp-Gen has shown remarkable zero-shot generalizability on other datasets. The source code is available at \href{https://github.com/CUHK-AIM-Group/Polyp-Gen}{https://github.com/CUHK-AIM-Group/Polyp-Gen}.

\end{abstract}


\section{INTRODUCTION}

\begin{figure}[t]
  \centering
\includegraphics[width=\linewidth]{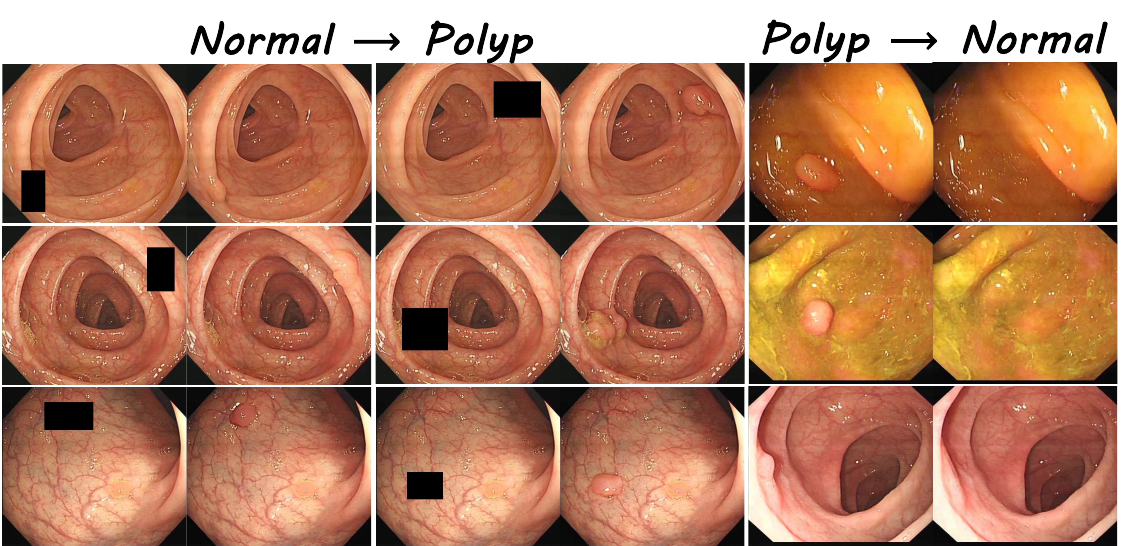}
   \caption{\textbf{Examples of the generated endoscopic images by our Polyp-Gen.} Existing endoscopic image generation methods failed to accurately generate the details of polyp boundary regions, while our Polyp-Gen framework achieves realistic and diverse endoscopic image generations.}
   \label{fig:first}
\end{figure}

Early detection of polyps in the lower gastrointestinal tract (GI) can significantly reduce the risk of developing life-threatening colorectal cancer. During this procedure, gastroenterologists typically need to identify polyps from lengthy videos, which is greatly reliant on their expertise and requires considerable labor. With the development of artificial intelligence and deep learning techniques, automated diagnostic systems (ADS) demonstrate significant potential in endoscopic examinations \cite{chen2023medical, pranet, chen2023surgical, liu2024deep}. The efficacy of ADS is highly dependent on the quantity and quality of the training data \cite{GAN_chen1, Endora}. However, due to high manual annotation costs and strict privacy protection requirements, acquiring high-quality data with realism and diversity poses a considerable challenge within the medical domain. 


Recent works \cite{PolypConnect, conditionalMaskDiff, PolypDDPM, thambawita2022singan, SemanticPolypGAN, simpleUnetpolyp, ControlPolypNet, PolypInpaint} proposed generation models for synthesizing endoscopic images to expand the polyp dataset, eliminating the laborious manual labeling. One category of approaches \cite{PolypConnect, PolypDDPM, SemanticPolypGAN, simpleUnetpolyp} directly generated polyp images with analogous structures based on the pixel-wise polyp mask, and another category of inpainting-based approaches \cite{ControlPolypNet, PolypInpaint} filled in polyps at the corresponding locations in normal images based on the polyp mask. Although these methods have achieved much progress in endoscopic image generation tasks, they still suffer from two main problems.

One problem is the failure to accurately generate the details of polyp boundary regions. Previous methods \cite{PolypDDPM, thambawita2022singan, simpleUnetpolyp} simply generated the polyp and normal regions according to the pixel masks, without taking into account their structural coherence. This approach limited the model's ability to accurately capture the structural context of polyp boundary regions, leading to endoscopic images with low resolution, color discrepancies, and structural inaccuracies. Another challenge is existing methods \cite{PolypDDPM, SemanticPolypGAN, ControlPolypNet} required to pre-specify the location where polyps will be generated. The pixel-wise masks used in the inpainting-based methods \cite{PolypConnect, ControlPolypNet} were typically derived directly from the training dataset without considering the plausibility of polyp locations during their generation. Alternatively, these masks were specified by experts who rely on their clinical expertise to identify plausible polyp locations and shapes, leading to high manual costs. These limitations constrain the practicality and effectiveness of endoscopic image generation models.


To address these limitations, we introduce \textbf{\textit{Polyp-Gen}}, the first full-automatic diffusion-based framework to create a wide range of realistic endoscopic images for dataset expansion. Specifically, we propose a spatial-aware diffusion training scheme to enhance the structural context of polyp boundary regions while preserving endoscopic global information, enabling the transformation between polyp images and normal images, as shown in Fig. \ref{fig:first}. In addition, to capture medical priors for the automatic identification of potential polyp locations, we devise a hierarchical retrieval-based sampling strategy with a mask proposer to match similar fine-grained spatial features. In this way, our generated images can achieve quality comparable to real images and enhance downstream diagnosis tasks. The contributions of our work can be summarised as follows:
\begin{itemize}
\item We propose Polyp-Gen, the first text-guided full-automatic diffusion-based endoscopic image generation framework.
\item We devise a boundary-enhanced
pseudo mask module with a lesion-guided loss to effectively learn the structural context of polyp boundary regions.
\item We introduce a hierarchical retrieval-based sampling strategy to adaptively determine potential polyp locations without relying on prior clinical knowledge.
\item We conduct comprehensive experiments to validate the realism and diversity of synthetic images. Our Polyp-Gen demonstrates strong zero-shot generation capabilities to other datasets.
\end{itemize}


\section{RELATED WORK}

\noindent\textbf{GAN-based models.} Early studies \cite{PolypConnect, thambawita2022singan, SemanticPolypGAN, simpleUnetpolyp} adopted Generative Adversarial Networks (GAN) \cite{GAN, GAN_lin1, GAN_chen2, GAN_lin2} for endoscopic image generation, especially polyp image generation. Among them, SinGAN-Seg \cite{thambawita2022singan} was trained using a single polyp image and corresponding mask to produce synthetic samples. However, this is a time-intensive procedure, and the resulting images closely resemble the original image. SemanticPolypGAN \cite{SemanticPolypGAN} utilized masked images as conditional input to generate polyp images, and Hemin \textit{et al.} \cite{simpleUnetpolyp} achieved a conversion between polyp image and normal image by a simple U-Net based conditional GAN. PolypConnect \cite{PolypConnect} devised a polyp inpainting pipeline by extracting and combining edge information. 

\noindent\textbf{Diffusion-based models.} Due to the convergence instability of GANs and their limited effectiveness in this area, diffusion models have emerged as a more popular alternative, offering the ability to produce more realistic images and aiding in tasks like text-to-image generation \cite{glide, imagen, stablediffusion, ukan}. Some recent works \cite{conditionalMaskDiff, PolypDDPM, ControlPolypNet} utilized diffusion models to produce polyp images. Polyp-DDPM \cite{PolypDDPM}, PolypInpainter \cite{PolypInpaint} and Roman \textit{et al.} \cite{conditionalMaskDiff}  both employed the pixel-wise mask of a polyp image to guide a conditional latent diffusion model for generating synthetic polyps. ControlPolypNet \cite{ControlPolypNet} utilized the ControlNet \cite{controlnet} architecture and diffusion model to generate polyp frames. However, these methods ic and diverse endoscopic images, making them ineffective for building ADS.


\section{METHODS}

\begin{figure*}[t]
  \centering
   \includegraphics[width=0.93\linewidth]{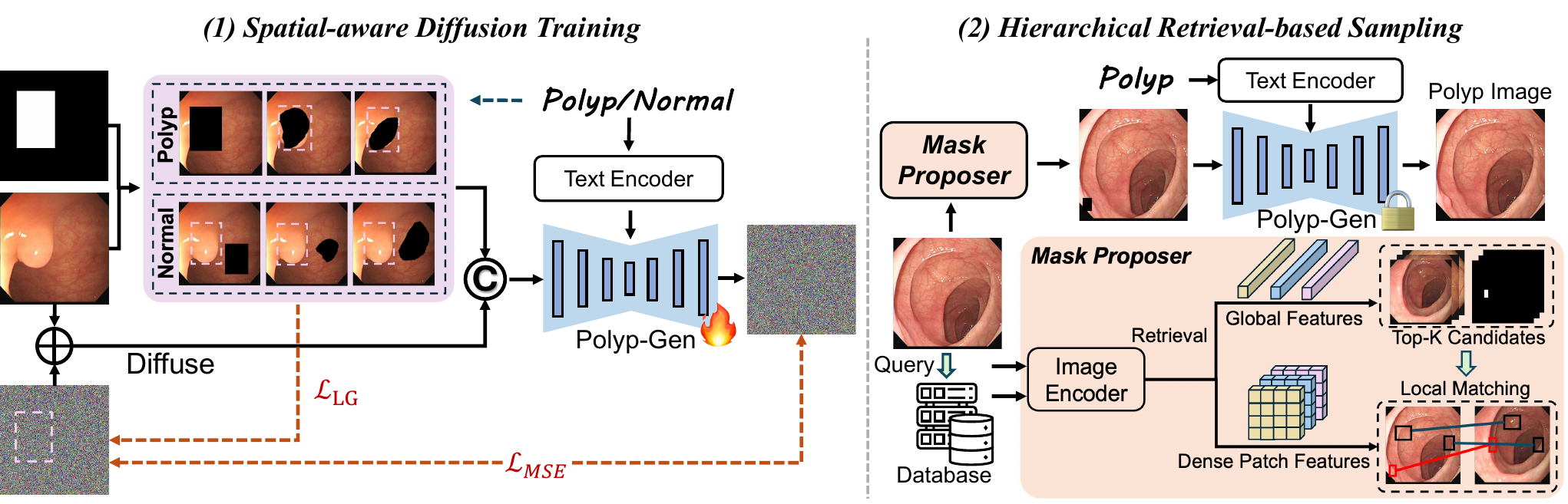}
   \caption{\textbf{Overview of Polyp-Gen.} (1) We devise a spatial-aware diffusion training scheme to enhance the structural context of polyp boundary regions while preserving endoscopic global information.
   (2) We introduce a hierarchical retrieval-based sampling strategy to adaptively determine potential polyp locations for polyp generation.}
   \label{fig:pipeline}
\end{figure*}

\begin{figure}[t]
  \centering
   \includegraphics[width=0.95\linewidth]{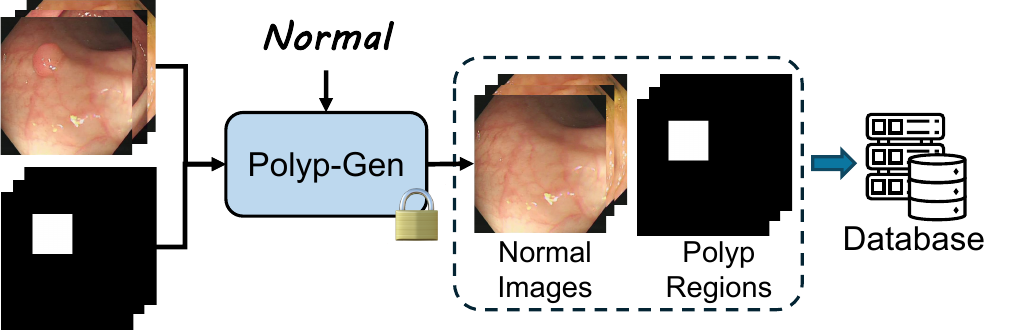}
   \caption{\textbf{Construction of database.} We utilize the trained Polyp-Gen to convert polyp images into normal images, thereby constructing a database for mask proposals.}
   \label{fig:database}
\end{figure}    

\subsection{Overview of Polyp-Gen}

As shown in Fig. \ref{fig:pipeline}, we propose Polyp-Gen, a full-automatic diffusion-based framework for the generation of realistic and diverse endoscopic images, both polyp and normal images. Given an endoscopic image $I\in R ^{ h \times w \times 3}$, a binary mask $M\in \{ 0, 1\} ^{h \times w}$, and a text prompt $P=\{\textit{Polyp},  \textit{Normal} \}$, our Polyp-Gen aims to generate the target image $I' = \theta(I, M, P)$ corresponding to the text while preserving background consistency. The mask value of 1 specifies the area to be inpainted, while 0 indicates that the context pixels should not be altered.

Polyp-Gen consists of two stages, including a spatial-aware diffusion training stage and a hierarchical retrieval-based sampling stage. In the training stage, we train the diffusion model with the boundary-enhanced pseudo mask module and the lesion-guided loss. Then we utilize the trained Polyp-Gen to convert polyp images into normal images, thereby constructing a database for mask proposals. In the sampling stage, we employ the hierarchical retrieval-based
sampling strategy to adaptively generate polyps within the original input non-polyp image.

\subsection{Stable diffusion model}
We implement our Polyp-Gen based on the Stable Diffusion (SD) \cite{stablediffusion}, which demonstrates superior ability in text-guided image generation tasks. The SD is a two-stage diffusion model that contains a variational autoencoder (VAE) and a U-Net denoiser. The forward and backward process of SD is performed in the latent space allowing for better control and manipulation of the generated content. Given an input image $I$, the VAE encoder extracts the latent representation of the image $z = E(I)\in \mathcal{R} ^{ \frac{h}{8} \times \frac{w}{8} \times 4}$. In the forward process, a noisy latent variable $z_t$ is obtained by introducing random noise $\epsilon \sim N(0, 1)$ at time step $t$, given by $z_t = \alpha_{t}z_{t-1} + \sigma_{t} \epsilon$, where $\alpha_t$ and $\sigma_t$ are coefficients that govern the noise schedule. Subsequently, the conditional diffusion model $\epsilon_\theta$ is optimized using the following objective:
\begin{equation}
L_{D} = \mathbb{E}_{z, c, \epsilon \sim N(0,1), t}\left\|\epsilon-\epsilon_\theta\left(z_t,  c, t\right)\right\|_2^2,
\label{eq:diff}
\end{equation}
where $c$ represents the additional conditions, including text and image prompts. 
In the backward process, given input noise $z_{t}$ sampled from a random Gaussian distribution, the learnable network $\epsilon_\theta$ estimates noise at each step $t$ conditioned on $c$. After a step-by-step denoising process, we obtain the latent representation $z_0$, which is then decoded by the VAE decoder $D$ to obtain the generated image $x=D(z_0)$. DDPM \cite{DDPM}, DDIM \cite{DDIM}, and DPM-Solver \cite{DPMSolver} are frequently used samplers in the denoising process.

\subsection{Spatial-aware diffusion training}

In the training stage, to enhance the generation of the polyp boundary region and increase the diversity of the generated images, we first introduce a Boundary-enhanced Pseudo Mask (BPM) module.
Specifically, given the prompt $P=\textit{Polyp}$, we utilize the bounding box $B$ rather than the pixel-wise mask $M$ of the polyp image. Then we randomly generate inscribed convex polygons from $B$ as augmented masks to form a training sample pair. In this way, the model can explicitly and accurately capture the structural context of the polyp boundary, thereby generating more realistic polyp images. On the other hand, when $P=\textit{Normal}$, the training sample pairs are composed in two ways: (1) a non-polyp image with a random mask; (2) a polyp image with a random mask located outside the bounding box $B$. Hence, the model is trained to complete polyp or normal regions in specified areas, guided by text prompt $P$. With a binary mask $M$ and a masked image $I^{m} = (1-M) \odot I$, the latent representation of the masked image is obtained $z^{m}=E(I^{m})$, and the mask $M$ is resized to $m$ to match the size of the latent representation. The conditional diffusion model $\epsilon_\theta$ is optimized towards the following objective:
\begin{equation}
L_{MSE} = \mathbb{E}_{z, c, \epsilon \sim N(0,1), t}\left\|\epsilon-\epsilon_\theta\left( \left[z_t, z^{m}, m\right], c, t\right)\right\|_2^2,
\label{eq:diff2}
\end{equation}
In this way, the model $\epsilon_\theta$ is trained using the mean square error (MSE) loss between the predicted noise and the actual noise added at each step. However, MSE loss is calculated by averaging the values across all pixels of an entire image, which fails to focus on the crucial lesion areas.

To tackle this issue, we encourage the model to emphasize the lesion regions by devising a lesion-guided loss $L_{LG}$ that selectively amplifies the importance of the target areas based on the provided mask $m$.
\begin{equation}
L_{LG} = \mathbb{E}_{z, c, \epsilon \sim N(0,1), t}\left\|m\odot\epsilon-m\odot\epsilon_\theta\left( \left[z_t, z^{m}, m\right], c, t\right)\right\|_2^2
\label{eq:lgloss}
\end{equation}
The $L_{LG}$ guides the model to emphasize the importance of the masked region, allowing the model to flexibly adjust the attention area, thus enhancing the generation quality. Hence, the final loss function is calculated as follows:
\begin{equation}
L= L_{MSE} + \lambda L_{LG},
\end{equation}
where $\lambda$ is a weighting factor set as 0.5.

Compared with using a pixel-wise mask, the spatial-aware diffusion training scheme facilitates a more accurate reconstruction of polyp boundary regions and provides enhanced flexibility. Consequently, Polyp-Gen can generate more realistic and diverse endoscopic images, especially polyp images.

\subsection{Hierarchical retrieval-based sampling}

To enable the model to perceive medical priors related to polyp locations and thereby generate polyps in anatomically plausible positions, we propose a hierarchical retrieval-based sampling strategy that adaptively identifies potential polyp regions based on the input image.

During the sampling process, we introduce a simple and efficient retrieval-based mask proposer (RMP) to identify and match similar structures in a database, with a high likelihood of these structures being polyps. The construction of the database is shown in Fig. \ref{fig:database}. We employ a trained Polyp-Gen model to convert the polyp image $I_P$ with corresponding mask $M_P$ to normal image $I_N=\theta(I_P, M_P, \textit{Normal})$, and then build the database $\mathcal{D}=\{I_N, M_P\}$. The database enables a non-polyp image to serve as the query image for retrieving images with similar location, shape, and structure. 

Specifically, we devise a two-stage retrieval pipeline to obtain potential polyp regions using the pre-trained image encoder DINOv2 \cite{dinov2}, which has shown an excellent ability to match similar semantic patch-level features across domains \cite{zhang2024tale, Roma}. We discard the class token and reshape the patch tokens as dense local features $f^l$, and calculate the global feature by mean pooling $f^g = \textit{MeanPooling}(f^l)$. We first compute the $L2$ distance to perform the similarity search in the global feature space over the database to get the top-K most similar candidate images. Afterward, we conduct a local feature matching between the query image $I_q$ and the candidate images $I_c \in \mathcal{D}$. We then calculate the nearest neighbor matches $\mathcal{M}$ for patches in $I^q$ with the masked regions of $I_c$:
\begin{equation}
\mathcal{M} = \{p: p=\mathop{\arg\min}\limits_{u} d(u, v), I^{v}_{c} \in M_P \odot I_{c}\},
\end{equation}
where the $L2$ distance between $j$-th patch in $I_{q}$ and $k$-th patch in $I_{c}$:
\begin{equation}
d(j, k)=\Vert f_{q}^{l}(j), f_{c}^{l}(k)\Vert_2,
\end{equation}
where $j, k \in \{1, 2, \cdots, N\}$, $N=\frac{H}{P} \times \frac{W}{P}$ denotes the number of patches, and $P$ is the patch size. The $p$-th  patch corresponds to the $(h, w)$ coordinates in the query image space. To further obtain similar local structures from $\mathcal{M}$, we employ the density-based spatial clustering of applications with noise (DBSCAN) \cite{DBSCAN} algorithm to cluster points that exhibit high proximity. The radius $\epsilon_r$ is set as $2\times P+1$. We then select the cluster with the maximum number of points and obtain its enclosing rectangle as the mask proposal $M_q$. The position of the mask, representing the most similar local structure to the potential polyp region in the reference image, also indicates the most plausible location for polyp generation in the query image. After obtaining the mask proposal $M_q$, the trained Polyp-Gen can generate polyp images $I'_P=\theta (I_q, M_q, \textit{Polyp})$. 

Therefore, the retrieval-based mask proposer facilitates the Polyp-Gen in recognizing medical prior knowledge about polyp locations, enabling it to produce polyps in anatomically realistic positions. Furthermore, since a non-polyp image may contain multiple mask proposals, this method also enhances the diversity of the generated images.

\section{EXPERIMENTS}

\subsection{Experimental setup}
\noindent\textbf{Datasets.} Our experiments use public dataset LDPolypVideo \cite{LDPolypvideo} for Polyp-Gen training. This dataset contains 160 colonoscopy videos and 40,266 frames with polyp annotations, including 33,884 polyp frames and 61 non-polyp videos. We filter out some low-quality images with blurry, reflective, and ghosting effects, and finally select 55,883 samples including 29,640 polyp frames and 26,243 non-polyp frames. We randomly sample 2,000 images (1,000 polyp frames and 1,000 non-polyp frames) as the test set, with the remaining images used as the training set. The image resolution is $560\times480$ and all polyp frames have bounding box annotations. Moreover, we use Kvasir-Seg \cite{Kvasirseg} to validate the performance of our Polyp-Gen framework.

\noindent\textbf{Implementation details.} Polyp-Gen is trained based on the pre-trained parameters of the SD-inpainting model \cite{stablediffusion}. The U-Net has 9 input channels (4 for the noisy latents, 4 for the encoded masked-image latents, and 1 for the mask itself). We use the weights of SD VAE as it can achieve robust image reconstruction capabilities even for medical images. CLIP \cite{CLIP} is utilized as the text encoder to extract embeddings for text-guided generations. We employ the AdamW optimizer with a constant learning rate of $1\times10^{-5}$. The implementation utilizes PyTorch and Diffusers library \cite{diffusers} for execution. The training process is conducted on an NVIDIA 4090 with a batch size of 2, and the number of gradient accumulation steps is 4. A 50-step DDIM \cite{DDIM} sampler is used in the inference stage to generate images. 

\noindent\textbf{Evaluation metrics.}
We use two evaluation metrics to evaluate the realism and diversity of generated endoscopic images: Frechet Inception Distance (FID) \cite{FID} and Inception Score (IS) \cite{IS}. FID is a metric used to quantify the dissimilarity in distribution between generated images and real images. A lower FID value generally signifies a greater resemblance between the generated and real images. IS quantifies the quality and diversity of generated images through the exponential average of image class clarity and variety. Additionally, we used task-specific detection metrics, including Average Precision (AP),  Precision, Recall, and F1-Score. The AP is the area under the curve (AUC) of the Precision $\times$ Recall. The F1-Score calculates the harmonic weight of the precision and recall.

\begin{figure}[b]
  \centering
\includegraphics[width=0.98\linewidth]{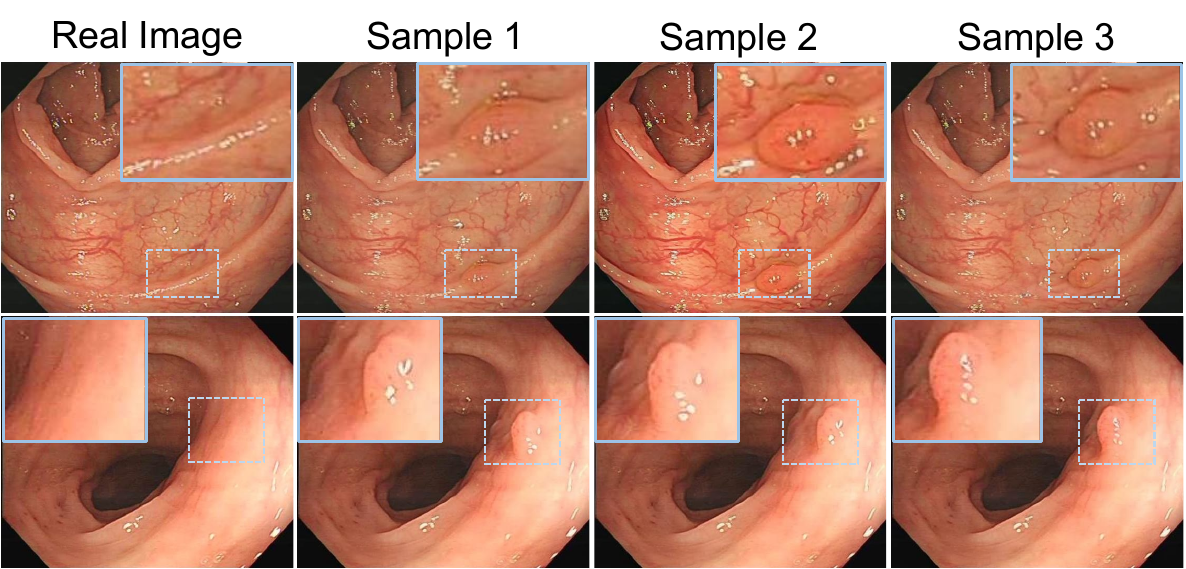}
   \caption{\textbf{Showcases of Polyp-Gen.} The generated images exhibit remarkable realism and diversity.}
   \label{fig:diversity}
\end{figure}

\subsection{Image synthesis performance}
\begin{figure}[h]
  \centering
\includegraphics[width=0.95\linewidth]{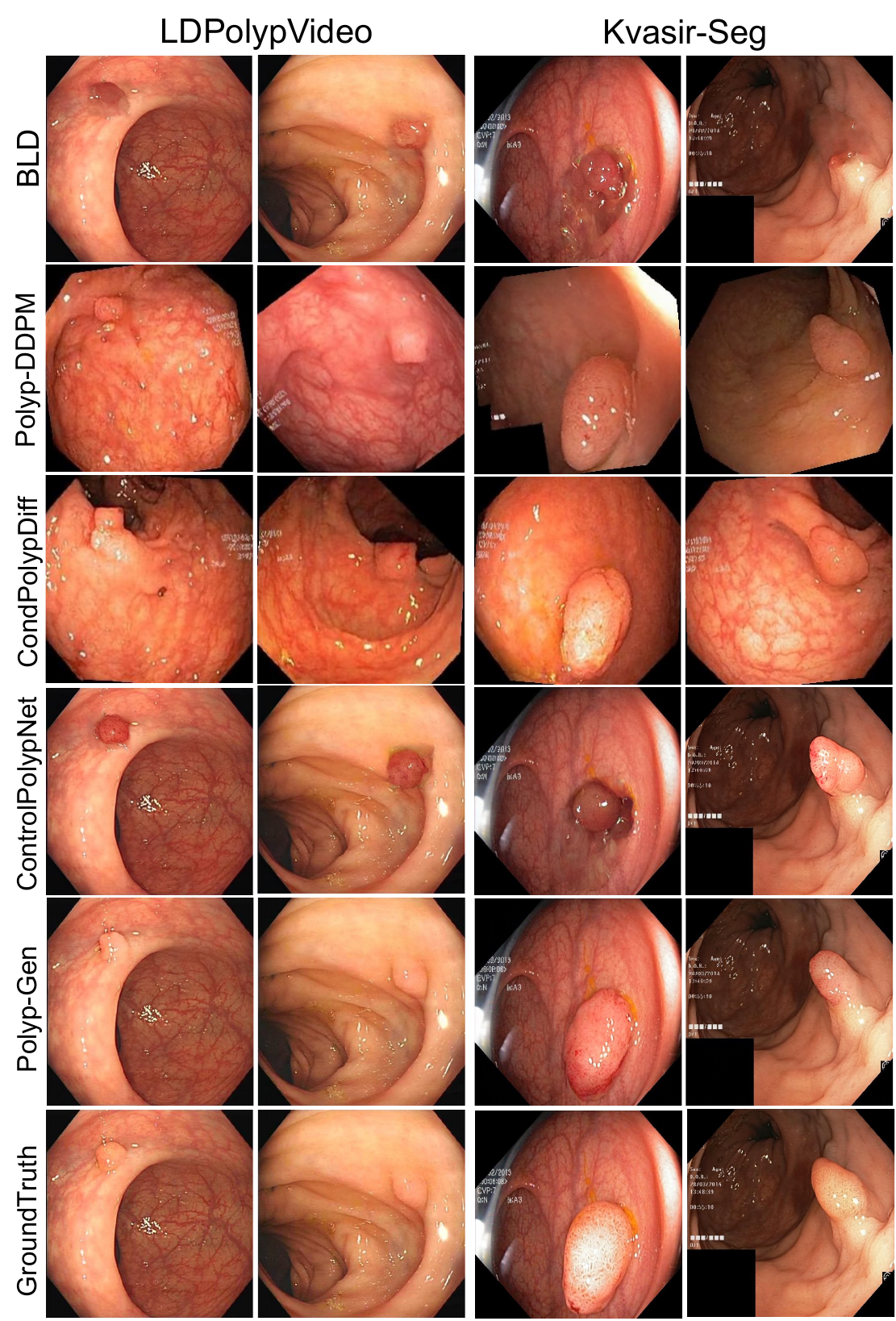}
   \caption{\textbf{Qualitative comparison} of polyp image generation on LDPolypVideo and Kvasir-Seg datasets.}
   \label{fig:compare}
\end{figure}

We comprehensively compare our Polyp-Gen framework and state-of-the-art endoscopic image generation methods \cite{conditionalMaskDiff, PolypDDPM, blendedDiffusion, ControlPolypNet}. Among them, Polyp-DDPM \cite{PolypDDPM} and Conditional polyp diffusion \cite{conditionalMaskDiff} employ the pixel-wise masks as conditional inputs, while Blended Latent diffusion (BLD) \cite{blendedDiffusion},  ControlPolypNet \cite{ControlPolypNet}, and our Polyp-Gen utilize the masked negative images as inputs. Each model generates four samples from each image in the test set, resulting in a total of 8,000 images. As shown in Table \ref{tab:fidis}, our Polyp-Gen achieves the lowest FID score of 13.817 on the LDPolypVideo dataset. This indicates that the images generated by Polyp-Gen are of superior quality and exhibit a greater level of realism. Moreover, our Polyp-Gen demonstrates the highest diversity with an IS of 3.454. Fig. \ref{fig:diversity} showcases the polyp images generated by Polyp-Gen. Compared with inpainting-based methods, Polyp-DDPM and CondPolypDiff exhibit relatively poorer performance in terms of image generation quality and diversity as they rely on pixel-wise masks as conditional inputs. By prioritizing the fulfillment of these mask conditions, these models are susceptible to overlooking vital image features, ultimately leading to a loss of detail or the production of unnatural appearances.
Additionally, considering the substantial difference in the size of our training set (55,883 images) and the Kvasir-Seg dataset (1,000 images), we conducted a zero-shot generation test on Kvasir-Seg. As shown in Table \ref{tab:fidis}, polyp images generated by Polyp-Gen exhibit an FID of 61.862 and an IS of 2.457. Fig. \ref{fig:compare} shows the qualitative comparison between different models. These results illustrate the remarkable generation capability of Polyp-Gen without any training and demonstrate its potential for generalization to other datasets.

\begin{table}[t]
\setlength{\tabcolsep}{2.5mm}
\centering
\caption{Comparison of different endoscopic image generation models on LDPolypVideo and Kvasir-Seg datasets. The best value of each column is highlighted in \textbf{bold}.}
\label{tab:fidis}
\begin{tabular}{ccccc}
\toprule
\multirow{2}{*}{\textbf{Method}} & \multicolumn{2}{c}{\textbf{LDPolypVideo}} & \multicolumn{2}{c}{\textbf{Kvasir-Seg}} \\
\cmidrule(lr){2-3} \cmidrule(lr){4-5}
& \textbf{FID}$\downarrow$ & \textbf{IS}$\uparrow$ & \textbf{FID}$\downarrow$  & \textbf{IS}$\uparrow$ \\
\midrule
BLD \cite{blendedDiffusion}&  20.252 & 2.978 & 92.755 & 2.087  \\
Polyp-DDPM \cite{PolypDDPM}   & 32.017 & 2.721 & 149.683 & 1.795 \\
CondPolypDiff \cite{conditionalMaskDiff}  & 29.536 & 2.665 & 127.034 &  1.864  \\
ControlPolypNet \cite{ControlPolypNet} & 17.230 & 3.126 & 67.811 & 2.193  \\
\midrule
Polyp-Gen & \textbf{13.817} & \textbf{3.454} & \textbf{61.862}  &  \textbf{2.457} \\
\bottomrule
\end{tabular}
\end{table}

\begin{figure}[h]
  \centering
\includegraphics[width=0.92\linewidth]{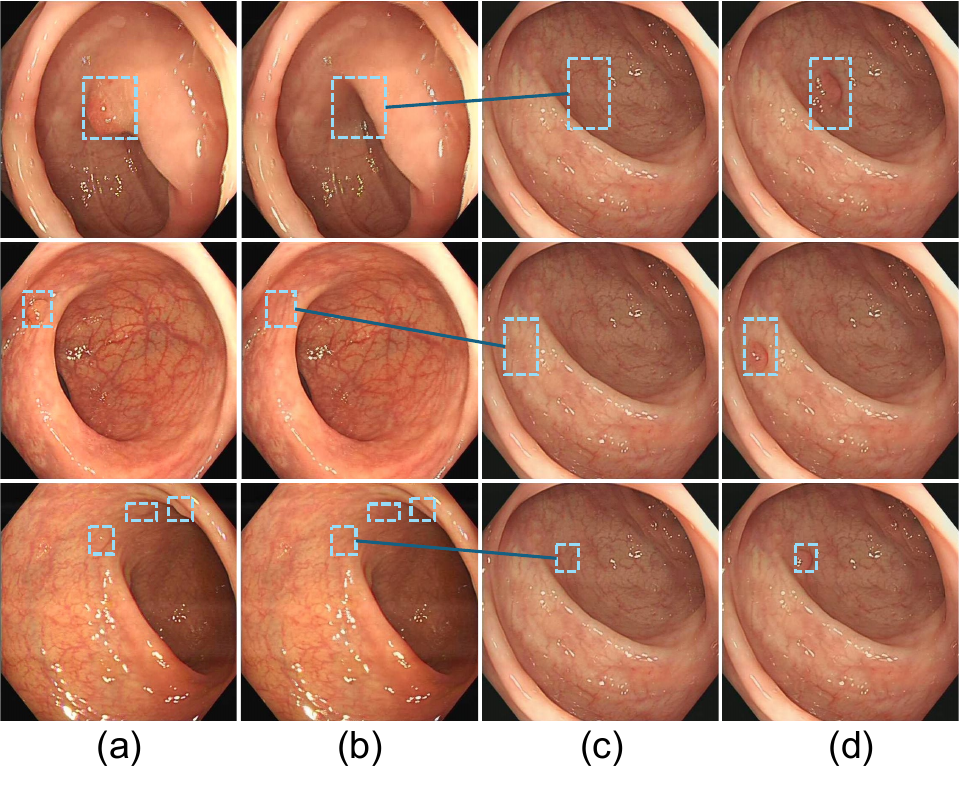}
   \caption{\textbf{Showcases of the retrieval-based mask proposer.} (a) Origin polyp images. (b) Top-3 reference non-polyp images generated from (a). (c) The non-polyp query image. (d) Generated polyp images by corresponding mask regions.}
   \label{fig:MaskPropose}
\end{figure}

Moreover, we also showcase the generation with retrieval-based mask proposer in Fig. \ref{fig:MaskPropose}. The results show that the mask proposer exhibits excellent fine-grained matching capability and enables the model to generate diverse images, thereby further demonstrating the effectiveness of our Polyp-Gen framework.

\begin{table*}[htb]
\setlength{\tabcolsep}{1.5mm}
\centering
\caption{Performance of the CenterNet \cite{CenterNet}, DINO \cite{DINO}, and YOLOv8 \cite{YOLOv8} models in the downstream task of polyp detection. The best results are highlighted in \textbf{bold}.}
\label{tab:detection}
\begin{tabular}{c cccc cccc cccc}
\toprule
\multirow{2}{*}{Method} & \multicolumn{4}{c}{\textbf{CenterNet} \cite{CenterNet}} & \multicolumn{4}{c}{\textbf{DINO} \cite{DINO}} &\multicolumn{4}{c}{\textbf{YOLOv8} \cite{YOLOv8}}\\
\cmidrule(lr){2-5} \cmidrule(lr){6-9} \cmidrule(lr){10-13} &
 \textbf{AP} & \textbf{Precision} & \textbf{Recall} & \textbf{F1-Score} &  \textbf{AP} &  \textbf{Precision} & \textbf{Recall} & \textbf{F1-Score} & \textbf{AP} & \textbf{Precision} & \textbf{Recall} & \textbf{F1-Score} \\
\midrule
Real Only & 56.74 & 69.45 & 43.17 & 53.24 &   62.42 & 70.47 & 48.50 & 57.46 &  64.37 & 70.72 & 50.13 & 58.67  \\
BLD \cite{blendedDiffusion} & 58.18 & 69.87 & 44.14 & 54.10 &   64.27 & 73.35 & 48.82 & 58.62 & 65.45 &71.84 & 52.58& 60.72 \\
Polyp-DDPM \cite{PolypDDPM} & 55.96 & \textbf{73.04} & 41.71 & 53.09 &  63.94 & 71.65 & 49.52 & 58.56   & 65.77 & 74.42 & 50.38 & 60.08\\
CondPolypDiff \cite{conditionalMaskDiff} & 56.18 & 71.03 & 42.84 & 53.45 &  64.68 & 72.26 & 50.71 & 59.60   & 66.49 & 75.29 & 51.91 & 61.45\\
ControlPolypNet \cite{ControlPolypNet} & 60.73 & 70.52 & 46.39 & 55.96    & 65.17 & 70.92 & 51.03& 59.35   & 67.21 & 71.73 & \textbf{56.21} & 63.03\\
\midrule
Polyp-Gen  & \textbf{64.22} & 72.08 & \textbf{50.56} & \textbf{59.43} & \textbf{67.35} & \textbf{76.68} & \textbf{53.92} & \textbf{63.32}  & \textbf{69.54} & \textbf{78.47} & 55.82 & \textbf{65.23} \\
\bottomrule
\end{tabular}
\end{table*}

\subsection{Downstream task evaluation}
The generated polyp images can be directly employed in training polyp detection models without requiring mask annotations. Additionally, a pixel-wise mask can be acquired through semi-supervised segmentation models like Segment Anything (SAM) \cite{semi_li2, SAM, semi_li1}. This is crucial as it reduces the manual labeling costs associated with expert annotation of masks in medical data.

To further validate the effectiveness of generated images, we conduct comprehensive experiments on polyp detection tasks utilizing three state-of-the-art detection models, CenterNet \cite{CenterNet}, DINO \cite{DINO}, and YOLOv8 \cite{YOLOv8}. We randomly select 10,000 real images as a training set and 2,000 images as a test set. Additionally, we incorporate 5,000 synthetic images into the training set to evaluate the impact of generated data. As shown in Table \ref{tab:detection}, the performance of the three detection models is significantly improved after incorporating the generated images. Specifically, with the CenterNet model, Polyp-Gen achieves an AP of 64.22\%, which is 3.49\% higher than the second-best method ControlPolypNet, 7.48\% higher than training with real images only. This trend continues with the DINO and YOLOv8 models, where Polyp-Gen consistently outperforms other methods. For instance, with the DINO model, Polyp-Gen achieves an F1-Score of 63.32\%, surpassing ControlPolypNet by 3.97\% and the real image-only baseline by 5.86\%. These results underscore the effectiveness of Polyp-Gen in leveraging synthetic data across various architectures, demonstrating its robustness and ultimate advantage in polyp detection tasks.

\subsection{Ablation study}
\begin{table}[h]
\setlength{\tabcolsep}{2.5mm}
\centering
\caption{\textbf{Ablation study.} BPM denotes the boundary-enhanced pseudo mask module. LG-loss denotes the lesion-guided loss. RMP denotes the retrieval-based mask proposer.}
\label{tab:ablation}
\begin{tabular} {ccc|cc}
\toprule
 \textbf{BPM} & \textbf{LG-loss} & \textbf{RMP} & \textbf{FID}$\downarrow$ & \textbf{IS}$\uparrow$ \\
\midrule
   &  &     &            16.562 & 3.174  \\
\ding{51} &     &      & 14.836 & 3.379  \\
& \ding{51}      &     & 15.118 & 3.206  \\
\ding{51} & \ding{51} & & \textbf{13.817} & 3.454\\
\midrule
\ding{51} & \ding{51} & \ding{51}  & 13.924 & \textbf{3.586}  \\
\bottomrule
\end{tabular}
\end{table}

As shown in Table \ref{tab:ablation}, we perform a series of ablation experiments to validate the effectiveness of the proposed modules on the LDPolypVideo dataset. Compared with the vanilla baseline, our framework with the BPM module gains a 6.45\% (0.205) increase in IS score, and a 10.4\% (1.726) increase in FID score. This confirms that the specified bounding boxes help the model learn to finely delineate polyp boundaries, which enhances the realism and diversity of the generated images.
Moreover, our framework obtains an 8.66\% (1.434) increase in FID when adding the lesion-guided loss in Polyp-Gen. The advantage is mainly attributed to its ability to emphasize the model's focus on generating critical regions. Additionally, 
we evaluate the effectiveness of the mask proposer. The results indicate a slight reduction in FID, and an improvement of 0.132 in IS, suggesting an enhancement in the diversity of the generated images. In this way, the proposed Polyp-Gen benefits from these tailored designs, resulting in the performance advantage in generating realistic and diverse endoscopic images.

\begin{table}[t]
\setlength{\tabcolsep}{1.5mm}
\centering
\caption{\textbf{Reader study.} Two gastroenterologists (G1 and G2) are tasked to label images as real or synthetic (100 real images, $100\times3$ synthetic images generated by 3 models). The table shows the percentage of images being identified as real.}
\label{tab:user}
\begin{tabular}{c|c|ccc}
\toprule
& \textbf{Real} & \textbf{Polyp-DDPM} & \textbf{ControlPolypNet}  & \textbf{Polyp-Gen} \\
\midrule
G1 &  92.0\% & 60.0\% & 82.0\% & \textbf{90.0\%}  \\
G2 &  89.0\% & 32.0\% & 67.0\% & \textbf{82.0\%}  \\
\bottomrule
\end{tabular}
\end{table}

\subsection{Reader study}
Following previous works \cite{DiffTumor, Dreambooth}, we involve experts in the field of endoscopy to ensure a thorough evaluation of model effectiveness. Specifically, we generate 100 polyp images for each model, including BLD, ControlPolypNet, and our Polyp-Gen. We then randomly sample 100 real images from the test set. A total of 400 images (synthesis and real) are shuffled and presented to two experienced gastroenterologists (G1 and G2) who evaluate each image as real or fake without being informed of the proportion of real images included. Table \ref{tab:user} shows the percentage of images being identified as real. The results reveal a strong similarity between the synthetic data and real polyps, resulting in the misidentification of most synthetic polyps as real ones. R1 misclassifies 90\% of the images generated by Polyp-Gen as real, while the recall of real images is 92\%. As for R2, who has more experience, has better recognition accuracy for all generated images. However, 82\% of the images generated by Polyp-Gen are still judged as real images, significantly higher than the 67\% of ControlPolypNet and the 32\% of Polyp-DDPM. The results of the reader study indicate that the generated images can attain a quality equivalent to real images, implying that our Polyp-Gen has the capability to produce highly realistic endoscopic images.
\section{CONCLUSION}
In this work, we propose Polyp-Gen, the first full-automatic diffusion-based framework for endoscopic image generation. Specifically, we devise the spatial-aware diffusion training scheme with a lesion-guided loss to enhance the structural context of polyp boundary regions while preserving endoscopic global information. Moreover, Polyp-Gen leverages a hierarchical retrieval-based sampling strategy that adaptively identifies potential polyp locations, enabling the generation of realistic images without needing prior clinical knowledge. We conduct extensive experiments to validate the Polyp-Gen framework, which reveals remarkable advantages in generation quality and zero-shot generalizability. In summary, our Polyp-Gen has the potential to offer strong data support for the construction of automated diagnostic systems.

\section{Acknowledgment}
This work was supported by Innovation and Technology Commission - Innovation and Technology Fund ITS/229/22 and Hong Kong Research Grants Council (RGC) General Research Fund 14220622.








\bibliographystyle{IEEEtran}
\bibliography{IEEEabrv, refs}

\end{document}